\documentclass[a4paper,11pt]{article}
\usepackage{a4wide,epsfig}
\usepackage{amssymb,mathrsfs,amsmath}
\usepackage{hyperref}
\usepackage{amsmath,mathrsfs}
\usepackage{graphicx}
\usepackage{amsfonts}
\usepackage{amssymb}
\usepackage{tikz}
\usepackage{color}
\usepackage{makecell} 
\usepackage{authblk}

\newtheorem{theorem}{Theorem}

\newtheorem{corollary}[theorem]{Corollary}

\newtheorem{definition}[theorem]{Definition}

\newtheorem{lemma}[theorem]{Lemma}

\newtheorem{remark}[theorem]{Remark}

\begin{document}

\title{A Solvable Molecular Switch Model for Stable Temporal Information Processing}

\author[1]{Hendra I. Nurdin\thanks{\texttt{Email: h.nurdin@unsw.edu.au}}}
\author[2]{Christian A. Nijhuis\thanks{\texttt{Email: c.a.nijhuis@utwente.nl}}}
\affil[1]{School of Electrical Engineering and Telecommunications, University of New South Wales, Sydney, Australia}
\affil[2]{Hybrid Materials for Opto-Electronics Group, Department of Molecules and Materials, MESA+Institute for Nanotechnology, Molecules Center
and Center for Brain-Inspired Nano Systems, Faculty of Science and Technology, University of Twente, Enschede, The Netherlands}
\date{}                 

\maketitle

\abstract{This paper studies an input-driven one-state differential equation model initially developed for an experimentally demonstrated dynamic molecular switch that switches like synapses in the brain do. The linear-in-the-state and nonlinear-in-the-input model is exactly solvable, and it is shown that it also possesses mathematical properties of convergence and fading memory that enable stable processing of time-varying inputs by nonlinear dynamical systems. Thus, the model exhibits the co-existence of biologically-inspired behavior and desirable  mathematical properties for stable learning on sequential data. The results give theoretical support for the use of the dynamic molecular switches as computational units in deep cascaded/layered feedforward and recurrent architectures as well as other more general structures for neuromorphic computing. They could also inspire more general exactly solvable models that can be fitted to emulate arbitrary physical devices which can mimic brain-inspired behaviour and perform stable computation on input signals. }

\section{Main}
A challenge of current computing architectures and computing approaches is the vast, unsustainable increase in energy consumption \cite{deVries23,Rillig23}. Inspired by the energy efficiency of natural computing systems such as human brains, there is now a global effort to develop new neuromorphic (brain-inspired) hardware and algorithms that go well beyond traditional computing hardware based on the von Neumann architecture, see, e.g., \cite{Schuman22,Zolf24,ZLTCGZWY23,KCSGvdB24,CHGHS25}. Prominent models for neural-synaptic networks in the brain, e.g., \cite{KLSM20,BNS21,CBR24} and the references therein, are based on recurrent neural networks (RNNs) and variations thereof, including models such as the so-called Hopfield neural networks and firing-rate neural networks  \cite{CBR24}. The disadvantage of such models is that they require a large computational overhead when executed on conventional digital computers. 

Here we report on a recently discovered experimental dynamic molecular switch  \cite{Wang22} that represents  a new type of neuromorphic information processing unit.  The switch constitutes molecular hardware that can be reconfigured into different logic gates or electronic functions, and processes information similar to synapses, emulating basic functions based on spike-timing dependent plasticity such as Hebbian or Pavlovian learning \cite{Wang22,Wang24,Zhang24}.  A molecular switching model was developed  in \cite{Wang22} for this dynamic molecular switch, derived using methods of theoretical chemistry.  A key contribution of this paper is the elaboration and analysis of this remarkable model for the switch,  in that it simultaneously possesses the properties of 
\begin{enumerate}
\item[(i)] having only one state,

\item[(ii)] being exactly solvable, 

\item[(iii)] emulating synaptic behaviour, as demonstrated in both simulation and experiments \cite{Wang22}, and 

\item[(iv)] being able to perform stable processing of input signals.
\end{enumerate}
The last property will be elaborated further  below in the penultimate paragraph of this section. We are not aware, to the best of our knowledge, of other models that simultaneously exhibit these four properties.  We believe that our findings may lead to new relationships between learning processes in the brain and mathematical concepts of systems theory and machine learning.

RNNs are nonlinear dynamical systems represented by nonlinear ordinary differential equations (ODEs) that are not analytically solvable and challenging to analyze. Desirable properties of these models that have been of much interest from a computational perspective are their stability and robustness for processing input signals which represent external stimuli to a neural circuit. That is, roughly speaking, their ability to produce a consistent response to the same stimulus in the presence of disturbances and variations to the initial state of the network. Recent advances have successfully used methods from nonlinear systems theory to establish sufficient conditions for stable processing in classes of RNN models \cite{KLSM20,BNS21,CBR24}. However, RNNs are not the only kind of continuous-time differential equation models being considered. In \cite{Hasani22}, a linear-in-the state but non-linear-in-the-input model for a single cell in liquid time-constant (LTC) networks was studied as a model for neuronal interactions. Being a linear but time-dependent  model in the state, the time-dependency arising from the time-varying input signal to the model,  the ODE has an exact solution given by the variation of constants formula. 
This makes this type of model very attractive for modeling information processing in brain-inspired architectures.  However, numerically solving the ODE may still be challenging and the work \cite{Hasani22} shows that there is an approximate closed-form solution that is shown to be a tight approximation, which does not require numerically solving their ODE. This is particularly advantageous for efficient numerical simulations of large LTC models.

The  continuous-time  ODE model for the dynamic molecular switch in \cite{Wang22} is also linear-in-the-state but nonlinear-in-the input. However, it should be emphasized that the model in \cite{Wang22} and the single-cell model in \cite{Hasani22} behave differently. In particular, whereas the latter model converges asymptotically to a constant value $A$ (the constant synaptic reverse potential) regardless of the input signal $I$ (see \cite[Methods, Eq. (6)]{Hasani22}), the model in \cite{Wang22} produces a (time-varying) output signal that asymptotically only depends on the (time-varying) input signal. An overview of the dynamic molecular switch and its mathematical model will be given in \S \ref{sec:results} of this paper. It should also be mentioned that linear-in-the state and nonlinear-in-the-input models  have also recently appeared in other contexts, albeit in the discrete-time rather than continuous-time setting \cite{chen2019learning, chen2020temporal,ENS25b}. A discrete-time model induced by the continuous-time model in \cite{Wang22} is developed in \S \ref{subsec:properties-DT}. 

Another pertinent development is recent efforts to exploit the natural dynamics of physical nonlinear systems for analog computation, in particular for machine learning algorithms; see, e.g., \cite{Wright22,Momeni23,Yan24,Liang24}. Mathematical formalisms have then been developed to provide a theoretical foundation for using dynamical systems as temporal information processing systems in the context of the Volterra series \cite{boyd1985fading} and, more recently, in reservoir computing \cite{grigoryeva2018universal,grigoryeva2018echo,gonon2019reservoir} and recurrent neural networks \cite{MH19}. In particular, the properties of convergence \cite{pavlov2004convergent} and fading memory \cite{boyd1985fading} have been used with the Stone-Weierstrass Theorem \cite{Dieudonne69} to establish so-called liquid state machines, a form of reservoir computing, as a computational model for real-time computing in recurrent integrate-and-fire neurons in neural microcircuits \cite{MNM02,maass2004computational}. In the context of echo-state networks \cite{JH04}, the convergence property is referred to as the echo-state property. These properties ensure that the system gives a bounded state response to a bounded input signal and that asymptotically (i.e., in the infinite time limit) this response is determined solely by the input signal, independent  of the initial state of the system. Therefore, with these properties the system performs a stable and well-defined mapping of input signals to output signals. In the context of recurrent neural networks, such stability properties assist in the training of the parameters of recurrent neural networks via stochastic gradient descent \cite{MH19}. We show here that the dynamic molecular switch model possesses both the convergence and fading memory property.

Our results also pave the way for the use of dynamic molecular switches as potentially energy efficient computational units in deep cascaded/layered feedforward and recurrent architectures and other more general structures for neuromorphic computing. Another significant implication of our work is that this model could provide the basis for discovering other exactly solvable parameterized models for synaptic dynamics, which may be used as a generic class of effective models for data-driven modeling of physical artificial synapses.

\section{Results}
\label{sec:results}

 Recently, we have reported on a dynamic molecular switch (DMS) that constitutes molecular hardware which can be reconfigured into different logic gates or electronic functions, and processes information similar to synapses, emulating basic functions based on spike-timing dependent plasticity such as Hebbian or Pavlovian learning \cite{Wang22,Wang24,Zhang24}. Unlike static on/off switches that switch between the same distinct on/off value irrespective of how often or how fast they are switched, DMSs continuously evolve and change their switching probabilities depending on their switching history and switching speed. Importantly, the properties of these switches could be captured using  a differential equation model \cite{Wang22} that is commonly used to describe current flow across molecules (i.e., Landauer-Büttiker formalism and Marcus Theories). In the following sections we show that this ODE model is exactly solvable and has the convergence and fading memory properties, but first we give a brief description of the system here and introduce the mathematical model of the DMS in the next section (for the more interested reader we refer to ref. \cite{Wang24,Zhang24} for a complete description).  The DMS consists of a self-assembled monolayer immobilized between two electrodes. The self-assembled monolayer has  5,6,11,12,17,18-hexaazatrinaphthylene (HATNA) terminal groups which can undergo 6 electron transfer steps coupled with 6 proton transfer steps so that the molecules are always in a charge neutral state irrespective of their oxidation state. By applying a voltage across the electrode—HATNA—electrode structure, a current flows across the junction governed by quantum mechanical tunneling. This current depends on the energy-level alignment of the system and the redox state of the molecule, which can be modelled by the Landauer equation. Depending on the redox-state of the molecule, the tunneling current is low (off-state) or high (on-state). Switching between the on and off states requires proton coupling steps, which are modelled using Marcus rate equations. The dynamic properties of the switch arise from the very different time-scales at which the fast quantum tunneling rates \cite[Eq. (6)]{Wang22} and slow proton hopping events \cite[Eq. (8)]{Wang22} occur leading to time-dependent changes in switching probabilities.

\textbf{Notation.} $\mathbb{R}$  denotes the set of real numbers and $\mathbb{R}_-=(-\infty,0]$. $\mathbb{Z}$  denotes the set of all real integers and $\mathbb{Z}_-$ the set of all non-positive integers (including 0).  A signal (function of time) will be denoted by $V_{\cdot}$ where the subscript $\cdot$ is a placeholder for time. If a signal is clear from its context then it will be denoted simply as $V$ (without the subscript). 

\subsection{Mathematical model of the dynamic molecular switch}
\label{subsec:DMS-model}
An approximate mathematical model for the DMS was proposed in \cite{Wang22} based on the Marcus charge theory and a theoretical framework developed by Migliore and Nitzan \cite{MN13}. The model is a single state dynamical system with the state being the probability for the junction potential of the DMS to be in the on-state (non-protonated state) AB, denoted by $P^{ AB}_{\cdot}$. It is given by the ODE  
\begin{align}
\dot{P}^{AB}_t &= (1-P^{AB}_t) k_{01}(V_t) - P^{AB}_t k_{10}(V_t)\label{eq:DMS-ODE-1}
\end{align}
where $V_t$ is the {\em bias} or {\em input} voltage to the DMS at time $t$, 
\begin{align}
k_{01}(v) &= (1-\langle b_n \rangle^{AB}(v))
R_{PT,+1}(V_t)  + \langle b_n \rangle^{AB}(v) R_{PT,+0}(V_t) \notag \\ 
k_{10}(v) &= (1-\langle b_n \rangle^{\overline{AB}}(v))
R_{PT,-1}(V_t)  + \langle b_n \rangle^{\overline{AB}}(v) R_{PT,-0}(V_t) \notag \\
R_{PT,\pm s}&=\frac{\gamma}{2}\sqrt{\frac{\pi k_B T}{\lambda}} e^{-\frac{( \alpha_s(V_t) \pm \lambda)^2}{4 k_B T \gamma_s}},\; s \in \{0,1\}, \notag \\
\alpha_1(v) &= v-E_{PT}, \notag  \\
\alpha_0(v) &= v-E_{PT}-\chi, \notag \\
\gamma_1 &= \gamma, \notag \\
\gamma_0 &= \kappa \gamma, \notag 
\end{align}
 $\langle b_n \rangle^{AB/\overline{AB}}$ is the average  bridge population in the non-protonated ($AB$) and protonated  ($\overline{AB}$) state that is a function of the bias voltage $v$ (formulae for the bridge population are given below), $T$ is the DMS junction temperature,  $k_B$ is the Boltzmann constant expressed in units of eV/K ($k_B =8.6173  \times 10^{-5}$ eV/K), $\lambda$ is the reorganisation energy in response to protonation, $\gamma$ is the molecule-surroundings coupling parameter modulating the protonation process, and $E_{PT} $ represents an energy level associated to the reduction process. Note that although \cite[Supplementary Material, \S S.6]{Wang22}) allows two distinct $E_{PT}$ values, $E_{PT,+}$ and $E_{PT,-}$, here we take them to be equal. This is in line with the single value of $E_{PT}$ provided in \cite[Supplementary Material, \S S.6, Table S15]{Wang22}). In the above expressions, $\kappa$ and $\chi$ are two constants that have been experimentally estimated. The average bridge population $\langle b_n \rangle^{AB/\overline{AB}}$ is given by the integral:
\begin{align*}
\lefteqn{\langle b_n \rangle^{AB/\overline{AB}}(v)}\\ 
&= \frac{1}{\gamma^{AB/\overline{AB}}}\int_{-\infty}^{\infty} (\gamma_L^{AB/\overline{AB}}f_+(E)(v) + \gamma_R^{AB/\overline{AB}}f_-(E)(v))  D_{E^{AB/\overline{AB}}}^{AB/\overline{AB}}(E)(v) dE,
\end{align*}
where
\begin{align*}
D^{AB/\overline{AB}}_{E'}(E)(v)
&= \frac{\gamma^{AB/\overline{AB}}/(2\pi)}{(E-(E'+ (\eta^{AB/\overline{AB}}-1/2)v))^2 + (\gamma^{AB/\overline{AB}}/2)^2},
\end{align*}
\begin{align*}
f_{\pm}(E)(v) &= \frac{1}{1+\exp\left((E \pm v/2)/k_B T \right)}
\end{align*}
In the expressions above:
\begin{itemize}

\item $\gamma_L^{AB/\overline{AB}} \geq 0$ and $\gamma_R^{AB/\overline{AB}} \geq 0$ are the tunneling rates between molecules in the junction and the left (L) and right (R)  electrodes, respectively, in the nonprotonated ($AB$) and protonated ($\overline{AB}$) states.

\item In \cite[Supplementary Material, \S 6.3]{Wang22} (note that $E_{AB/\overline{AB}}$ are denoted therein by $\epsilon^{AB/\overline{AB}}$), the pair of parameters $\gamma_{L/R}^{AB/\overline{AB}}$ and $E_{AB/\overline{AB}}$ are related via $\gamma_{L/R}^{\overline{AB}}=\kappa \gamma_{L/R}^{AB}$ and $E_{\overline{AB}} = E_{AB} + \chi$, where $\kappa$ and $\chi$ are the  two parameters alluded to previously. 

\item $\gamma^{AB/\overline{AB}}  = \gamma_L^{AB/\overline{AB}} +\gamma_R^{AB/\overline{AB}}$. Therefore, it holds that
$$
\gamma^{\overline{AB}}=\kappa \gamma^{AB}.
$$

\item $\eta^{AB/\overline{AB}}$ is a voltage division parameter. Here we will assume the scenario where $\eta^{AB}=\eta^{\overline{AB}}=V_R/(V_L+V_R)$, where $V_R$ and $V_L$ are the potential drops between the HATNA molecule and the top and bottom electrodes, respectively.

\end{itemize}

The model presented here is more detailed than the one in \cite[Eq. (2) and Supplementary Material Eq. (S5)]{Wang22} that makes the simplification $k_{01} \approx R_{PT,+1}$ and $k_{10} \approx R_{PT,-1}$. However, all the numerical simulation results reported in \cite{Wang22} were generated according to the detailed model described above \cite{Nickle25}, which suitably adapts the approach of Migliore and Nitzan \cite{MN13}. Properties of the model will be given in this Results section.  

The average output current $\overline{I}_{\cdot}$ of the DMS is given by \cite[Eq. (1)]{Wang22}:
\begin{align}
\overline{I}_t &= I^{AB}(V_t) P^{AB}_t + I^{\overline{AB}}(V_t) P^{\overline{AB}}_t, \label{eq:junction-current}\\
I^{AB/\overline{AB}}(v) &= \frac{Nq}{2 \pi \hbar} \int_{\mathbb{R}\times \mathbb{R}} dE dE' D^{AB/\overline{AB}}_{E'}(E)(v)  G_{E_{AB/\overline{AB}}}^{AB/\overline{AB}}(E')\Gamma^{AB/\overline{AB}} \notag\\ &\qquad \times (f_-(E)(v)-f_+(E)(v)),
\end{align}
where $I^{AB}_{\cdot}$ ($I^{\overline{AB}}_{\cdot}$) is the current flowing in the DMS junction when it is in the protonated (non-protonated) state,  $P^{\overline{AB}}_{\cdot}=1-P^{AB}_{\cdot}$ is the probability of the junction being in the protonated (off) state, $N$ is the number of molecules in the junction, $\hbar$ is the reduced Planck's constant in units of eVs$^{-1}$( $\hbar = 6.5821 \times 10^{-16}$ eVs$^{-1}$) and $q$ is the electron charge ($q=1.60217663 \times 10^{-19}$ C), and
\begin{align*}
G_{E_{AB/\overline{AB}}}^{AB,\overline{AB}}(E') &= \frac{1}{\sqrt{2\pi}\sigma^{AB/\overline{AB}}} \exp\left(-\frac{(E'-E_{AB/\overline{AB}})^2}{2(\sigma^{AB/\overline{AB}})^2}\right),\\
\Gamma^{AB/\overline{AB}} &= \frac{\gamma_L^{AB/\overline{AB}} \gamma_R^{AB/\overline{AB}} }{\gamma_L^{AB/\overline{AB}} + \gamma_R^{AB/\overline{AB}}}.
\end{align*}
Since $\gamma_{L/R}^{\overline{AB}} = \kappa \gamma_{L/R}^{AB}$
note that it follows that 
$\Gamma^{AB/\overline{AB}}$ are related via
$$
\Gamma^{\overline{AB}} = \kappa \Gamma^{AB}.
$$

Let $A(v) = - (k_{01}(v) + k_{10}(v))$. Also, let $\Phi_{t,\tau}$ for any $\tau$ and $t \geq \tau$ be the two-time transition function of the DMS satisfying the ODE:
$$
\dot{\Phi}_{t,\tau} = A(V_t) \Phi_{t,\tau}, 
$$
with initial condition $\Phi_{\tau,\tau}=1$. The unique solution $\Phi_{t,\tau}$ to the ODE is given by the exponential:
$$
\Phi_{t,\tau} = e^{\int_{\tau}^t A(V_{s})ds},\; t \geq \tau.
$$

The ODE for $P^{AB}_{\cdot}$ can be expressed as:
\begin{align}
\dot{P}^{AB}_t &= k_{01}(V_t) - (k_{01}(V_t) + k_{10}(V_t)) P^{AB}_t.  \label{eq:DMS-ODE-2}
\end{align}
It is a time-varying linear ODE that has a unique solution given by
\begin{align}
P^{AB}_t = \Phi_{t,t_0} P^{AB}_{t_0} + \int_{t_0}^t \Phi_{t,\tau} k_{01}(V_{\tau})d\tau, \; t \geq t_0,\label{eq:P-solution}
\end{align}
where $P^{AB}_{t_0}$ is the  probability for the non-protonated (on) state at the initial time $t_0$. 

In the following the main results that establish properties of the DMS model will be stated without their proofs.  All proofs are collected together in the Methods section. The following lemma establishes that $P^{AB/\overline{AB}}_t \geq 0$ for $t \geq t_0$ whenever $P^{AB/\overline{AB}}_{t_0}\geq 0$.  

\begin{lemma}\label{lem:positivity}
Let $\gamma^{AB/\overline{AB}}>0$. Then $0 \leq \langle b_n^{AB/\overline{AB}} \rangle(v) \leq 1$ for all $v$ and $P^{AB/\overline{AB}}_t \geq 0$ for all $t \geq t_0$ if $P^{AB/\overline{AB}}_{t_0} \geq 0$.
\end{lemma}

\begin{remark}
Following from Lemma \ref{lem:positivity} it will be assumed throughout the remainder of the paper that $\gamma^{AB/\overline{AB}}>0$. 
\end{remark}

The corollary below then establishes that indeed $P^{AB}_t$ satisfying the ODE is a proper probability at each time $t$ and thus so is $P^{\overline{AB}}_t$:
\begin{corollary}
\label{cor:proper-PAB}
$0 \leq P^{AB}_t,P^{\overline{AB}}_t \leq 1$ for all $t \geq t_0$ if $0 \leq P^{AB}_{t_0},P^{\overline{AB}}_{t_0}  \leq 1$.
\end{corollary}

It should be emphasized that the model is linear in $P_{\cdot}^{AB}$ but there is a nonlinear dependency of $P_{\cdot}^{AB}$ on the bias voltage signal $V_{\cdot}$. Due to this non-linearity a different response from the DMS can be obtained for the same bias voltage by rescaling and displacing this voltage. That is, the DMS can give different responses for both $P^{AB}_{\cdot}$ and $I^{AB/\overline{AB}}_{\cdot}$ for displaced and scaled bias inputs $a+kV_{\cdot}$ with different real displacements $a$ and scaling factors $k$ (that can be negative). Thus for the same bias voltage $V_{\cdot{}}$ a different current response could be be elicited from the DMS simply by performing this affine transformation on the voltage before applying it to the DMS. In this way, a diversity of dynamical responses to the same input can generated, a richness that is desirable in the machine learning context. 

For the numerical examples that will be presented throughout the paper, we consider the DMS model used to generate Fig. 5g in \cite{Wang22}, the parameter values of which are given in \cite[Supplementary Material Table S15]{Wang22} and also in the text of \S 6.3 of the supplementary material. These  values are summarized in Table \ref{tab:parameters} below.

\begin{table}[h]
\caption{DMS Parameter Values} 
\label{tab:parameters}
\begin{center}
\begin{tabular}{|c|c|}\hline 
Parameter & Value  \\    \hline
 \hline
$\Gamma^{AB}$  \hbox{(meV)} & 0.01 \\    \hline
$\sigma$ & 0.01  \\ \hline
$\eta$  & 0.6  \\ \hline
$E_{AB}$ \hbox{(eV)} & 0.66 \\\hline
$E_{PT}$  \hbox{(eV)} & -0.513 \\ \hline
$\kappa$ & 5.44\\ \hline
$\chi$ \hbox{(eV)} & 2.1  \\ \hline
$\lambda$ \hbox{(eV)} & 1 \\ \hline
$\gamma$ \hbox{($s^{-1}$)} & 5.74 \\ \hline
$\gamma_L$ \hbox{(meV)} & 4 \\ \hline
$\gamma_R$ \hbox{(meV)} & 100.25\\ \hline
$N$ \hbox{(number of molecules)} & 150\\
\hline
\end{tabular}
\end{center}
\end{table}

\begin{figure}[h]
\centering
\includegraphics[scale=0.42]{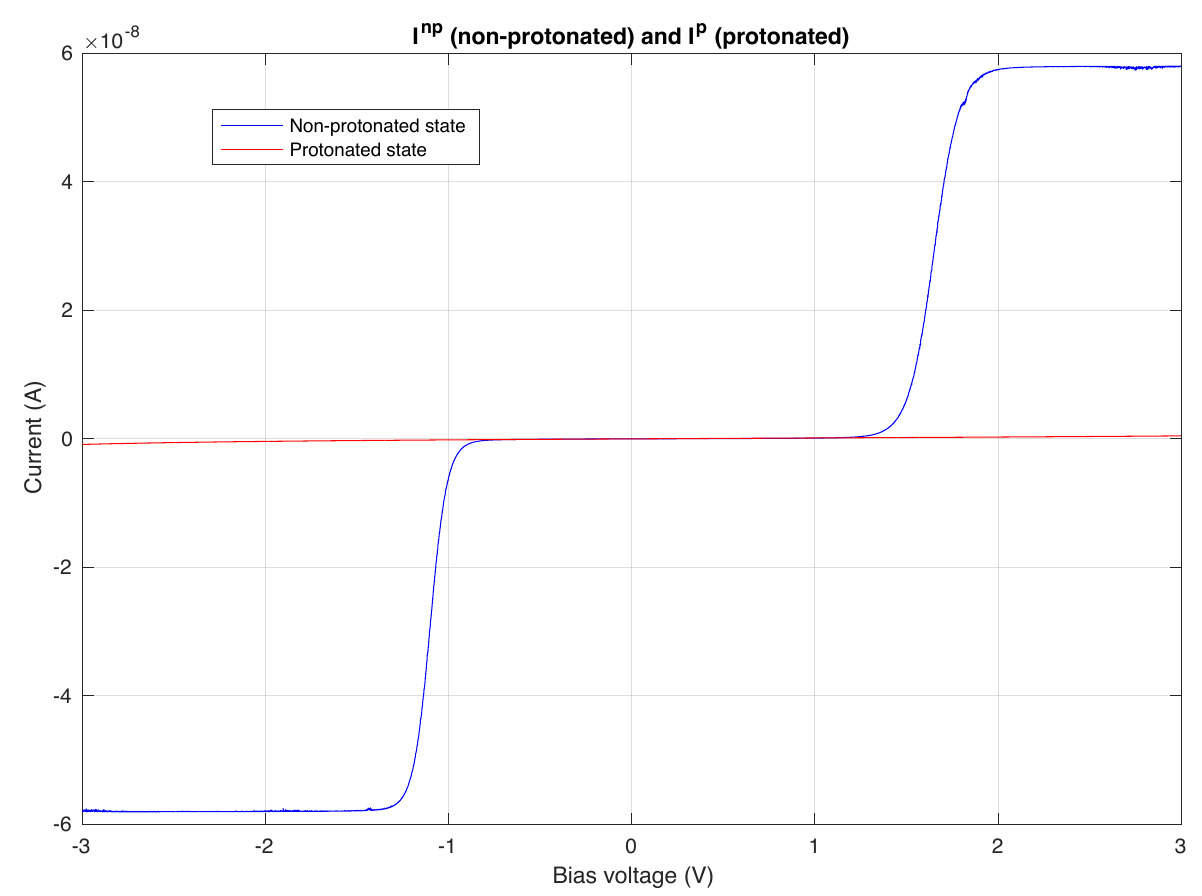}
\caption{Current response of the DMS vs input bias in the non-protonated (blue) and protonated state (red).}  
\label{fig:DMS-current-response}
\end{figure}

For a DMS with the parameters in Table \ref{tab:parameters}, Fig.~\ref{fig:DMS-current-response} shows its current response in the non-protonated ($I^{AB}$) and protonated ($I^{\overline{AB}}$)  states versus the bias voltage.  It can be seen that in the protonated state there negligible current flowing in the DMS for all values of the bias voltage. On the other hand, in the non-protonated state current only flows for sufficiently positive or negative values of the voltage and asymptotically settles to a constant value. On the other hand, if the magnitude of the voltage is not sufficiently large there is a negligible current response. 
 
The figure above was generated using an iterated numerical integration routine that partly uses Monte Carlo integration. First the integral  with respect to $E$ for a fixed value of $E'$ was computed for 500 independent random samples of $E'$ (which follows a normal distribution with mean $E_{AB/\overline{AB}}$ and variance $\sigma^2_{AB/\overline{AB}}$). Then the 500 integral values were averaged to get a numerical value for the double integral that appears in the expression for $I^{AB/\overline{AB}}$. This random sampling is the origin of the small fluctuations that is visible in the current plot for $I^{AB}$ for larger values of $v$. These fluctuations can be made smaller by using a larger number of random samples of $E'$. However, we did not feel it necessary to do so as the fluctuations are already relatively small and will not have a significant impact on the numerical calculations of the current values.

\subsection{Asymptotic and steady-state properties}
\label{subsec:asymptotic-ss}
In this section we will look at some asymptotic and steady-state properties of the DMS model. The first result is the asymptotic stability of the two-parameter transition function $\Phi_{t,t_0}$ as $t_0 \rightarrow -\infty$.

\begin{lemma}
\label{lem:stability}
Suppose that the input  (bias) voltage $V_t$ lies in an interval $[a,b]\subset \mathbb{R}$, with $a<b$ and both finite, for all $t$. Then $\mathop{\lim}_{t_0 \rightarrow -\infty} \Phi_{t,t_0} =0$.
\end{lemma}

Suppose that the bias voltage is constant, $V_t = v$ for all $t$ for some real constant $v$. Since the system is asymptotically stable by the above lemma, $P_t^{AB}$ will converge to a steady-state value $P_*^{AB}$ as $t \rightarrow \infty$ from any initial value. This steady-state value is given by the following corollary.

\begin{corollary}
\label{cor:steady-state}
The steady-state probability $P_*^{AB}$ for a constant bias voltage $v$ is given by 
\begin{align*}
P_*^{AB} &= \frac{k_{01}(v)}{k_{01}(v)+k_{10}(v)}.
\end{align*}
and the steady-state average junction current $\overline{I}_*$ is:
\begin{align*}
\overline{I}_* &= I^{AB}(v) P^{AB}_* + I^{\overline{AB}}(v) P^{\overline{AB}}_*. 
\end{align*}
\end{corollary}

\subsection{Convergence and fading memory properties of the dynamic molecular switch}
\label{subsec:properties-CT}
Here we will review the property of convergence \cite{PvdW16,pavlov2004convergent} and the notions of fading memory functional and filters \cite{boyd1985fading} for continuous-time nonlinear dynamical systems.  The latter originates from the study of Volterra filters as nonlinear maps from input signals to output signals. These notions  play an important role in the study of how dynamical systems can be used to approximately infer unknown relationships between input and output signals, which are central to problems of modeling unknown time-series and dynamical systems from data. For instance, input-output maps generated by Volterra filters can be used to learn fading memory maps from an input signal $\{u_t\}_{t \in \mathbb{R}}$ to an output signal  $\{y_t\}_{t \in \mathbb{R}}$. Roughly speaking, fading memory maps have the property that the dependence of $y_t$ on $u_{t-\tau}$ vanishes asymptotically as $\tau \rightarrow \infty$. In turn, fading memory maps can be induced by dynamical systems that have the so-called convergence property. That is, roughly speaking, the property of asymptotically forgetting its initial state for any initial time $t_0$. Since the DMS is a continuous-time system and can be operated with continuous-time signals, it is important to study its convergence and fading memory properties in continuous-time. Previously, the properties  of convergence and fading memory also played a central role in the proposal of liquid state machines as  a computational model for real-time computing in recurrent integrate-and-fire neurons in neural microcircuits \cite{MNM02,maass2004computational}.  The formal definitions are given as follows. 

\begin{definition}[\cite{PvdW16}]
\label{def:convergence-CT} An input-driven continuous-time dynamical system described by the ODE
$$
\dot{x}_t = f(x_t,u_t),
$$
with state $x_t \in \mathbb{R}^n$ and input signal $u_t \in \mathbb{R}^m$, and $f(z,u_t)$ locally Lipschitz in $z$ and piece-wise continuous in $t$ is said to be convergent with respect to the class of bounded piece-wise continuous inputs $u_{\cdot}$ on $\mathbb{R}$ if
\begin{enumerate}
\item  For each $u$, there is a unique solution $\bar{x}$ that  is defined and bounded for all $t \in \mathbb{R}$.

\item For each $u$,  a solution $x$ of the ODE that is initialized at $x(t_0)=x_0$ converges to the solution $\overline{x}$ in the sense that $\mathop{\lim}_{t_0 \rightarrow -\infty}\|x(t)-\bar{x}(t)\|=0$ for any $x_0$.
\end{enumerate}
 \end{definition}

In the following definition, let $\mathbb{V}(X,\mathcal{D})$ denote the class of signals $V_{\cdot}$ on the set $X \subseteq \mathbb{R}$ that are Lebesque measurable and take values in a compact set $\mathcal{D} \subset \mathbb{R}$, and let $w_{\cdot}$ be a real piece-wise continuous weighting function on $\mathbb{R}_-$ such that $w_t \in (0,1]$ for all $t \in \mathbb{R}_-$ and $\mathop{\lim}_{t\rightarrow -\infty} w_t=0$. Define the norm $\| \cdot \|_{w}$ for $V \in \mathbb{V}(\mathbb{R}_-,\mathcal{D})$ by $\|V\|_{w}=\mathop{\mathrm{ess} \sup}_{t \in \mathbb{R}_-}|w_t V_t|$ and the signal class $\mathscr{S}(\mathbb{R}_-,\mathcal{D})=\{V_{\cdot} \in \mathbb{V}(\mathbb{R}_-,\mathcal{D})  \mid \| V \|_{w} < \infty\}$. The class $\mathscr{S}(\mathbb{R}_-,\mathcal{D})$ together with the norm $\|\cdot\|_w$ defines a metric space $(\mathscr{S}(\mathbb{R}_-,\mathcal{D}),\|\cdot\|_{w})$ with the metric $d(V,V')=\|V-V'\|_{w}$.

\begin{definition}[\cite{boyd1985fading}] \label{def:CT-FM-functional}
A functional $F: \mathscr{S}(\mathbb{R}_-,\mathcal{D}) \rightarrow \mathbb{R}$ is said to be a fading memory functional (with respect to a weighting function $w$) if $F$ is a continuous functional on $(\mathscr{S}(\mathbb{R}_-,\mathcal{D}),\|\cdot\|_{w})$. 
\end{definition}

Let $\Pi_-$ denote the projection of a signal $V: \mathbb{R} \rightarrow \mathbb{R}$ onto its restriction  on $\mathbb{R}_-$, $(\Pi_-V)_t = V_t$ for all $t \leq 0$, and let $\sigma_t(V_{\cdot})$ be the shift operator acting on $V$ as $\sigma_t(V_s) = V_{s + t}$ for any $t \in \mathbb{R}$. Then we have the following definition. 

\begin{definition}[\cite{boyd1985fading}]
\label{def:CT-FM-filter}
A fading memory filter $U_F$ is a causal map from an input signal $V_{\cdot} \in \mathbb{V}(\mathbb{R},\mathcal{D})$ to an output signal $Y_{\cdot}$ induced by a fading memory functional $F$ as defined by
\begin{equation}
Y_t = U_F(V_{\cdot})_t=F(\Pi_-(\sigma_t(V_{\cdot}))
\end{equation}
for all $t \in \mathbb{R}$.
\end{definition}

\begin{theorem}
\label{thm:CT-FM}
Let $\mathcal{D}=[a,b]$ with $-\infty<a<b<\infty$ and let the weighting function $w$ satisfy 
$$ 
 \int_{-\infty}^{0}  \frac{e^{\nu \tau}}{w_{\tau}} d\tau < \infty,
$$
where $\nu = \min_{v \in \mathcal{D}}K(v)>0$ and $K=-A$. Then the  DMS model has the convergence property and induces the fading memory functional
$F(V) = \int_{-\infty}^0 \Phi_{0,\tau} k_{10}(V_{\tau})d\tau$ on $(\mathscr{S}(\mathbb{R}_-,\mathcal{D}),\|\cdot\|_{w})$
and the fading memory filter
$
U_F(V_{\cdot})_t =  F(\Pi_- \sigma_t(V)).
$
\end{theorem}

The condition on  the weighting function $w_{\cdot}$ means that it must decay to zero as $\tau \downarrow -\infty$ at a rate that is sufficiently slower than $e^{\nu \tau}$.  If $w_{\cdot}$ does not satisfy this condition it simply means that $|F(V)-F(V')|$ may not go to zero even if $\|V-V'\|_{w}$ goes to 0.  

\subsection{Discrete-Time Dynamics: Convergence and Fading Memory Properties}
\label{subsec:properties-DT}
Although the DMS is a continuous-time dynamical system, in applications such as time-series  modelling and system identification, they may be operated in discrete-time to process serial/sequential data (discrete-time signals); see, e.g., \cite{fan2008nonlinear,ljung2010perspectives,pavlov2008convergent,CN22,ENS25a}. This can be achieved by choosing the bias voltage to be piece-wise constant over a sampling time interval $T_s$. That is, set $V_t= \widetilde{V}_k$ for $kT_s \leq  t < (k+1) T_s$ for all integers $k$. Throughout this section  $\widetilde{\cdot}$ will be attached to variables to 
emphasize their association with a discrete-time system.

Let $\widetilde{P}_k^{AB} = P_{kT_s}^{AB}$. Then from \eqref{eq:DMS-ODE-2} we have the discrete-time dynamics:
\begin{align}
\widetilde{P}^{AB}_{k+1} = \widetilde{\Phi}_k \widetilde{P}^{AB}_{k} +  k_{01}(\widetilde{V}_k) \int_{kT_s}^{(k+1)T_s} \Phi_{t,\tau} d\tau, \label{eq:discrete-time}
\end{align}
where
$\widetilde{\Phi}_k = \exp\left(A(\widetilde{V}_{k}) T_s  \right)$. From the properties of $\Phi$, we have that $\widetilde{\Phi}_k$ satisfies $0 \leq \widetilde{\Phi}_k \leq e^{-K(\widetilde{V}_k) T_s} \leq e^{-\nu T_s}$, where as before $\nu=\min_{v \in \mathcal{D}} K(v)>0$ and $K=-A$. Since $\widetilde{\Phi}_k$ depends on $k$ through $\widetilde{V}_k$, it will also be written as $\widetilde{\Phi}_k = \widetilde{\Phi}_k(\widetilde{V}_k)$ to emphasize the dependence on $\widetilde{V}_k$, when necessary.

Notions of convergence and fading memory functionals reviewed in \S \ref{subsec:properties-CT} for continuous-time systems have discrete-time analogs that will be given below \cite{grigoryeva2018universal,tran2018convergence}. In the discrete-time reservoir computing context, the convergence property is often referred to as the {\em echo-state property} \cite{JH04,grigoryeva2018echo}. 

Recall that a function $\beta:[0,\infty) \times \mathbb{Z}_+ \rightarrow \mathbb{R}$ is a class $\mathcal{KL}$ function if $\beta(0,k) = 0$ for all $k$, is continuous and strictly increasing in the first argument, and is non-increasing in the second argument with $\mathop{\lim}_{k \rightarrow  \infty} \beta(x, k) = 0$ for all $x \in [0, \infty)$. 

\begin{definition}[\cite{pavlov2008convergent,tran2018convergence}]
\label{def:DT-convergence} An input-driven discrete-time dynamical system described by the difference equation
$$
x_{k+1} = f(x_k,u_k),\; k \in \mathbb{Z}
$$
with state $x_k \in \mathbb{R}^n$ and input signal $u_k \in \mathbb{R}^m$, and $f(x,w)$ being defined for all $x$ and $w$, is said to be (uniformly) convergent with respect to a class of bounded input sequences $u_{\cdot}$ on $\mathbb{Z}$ if
\begin{enumerate}

\item  For each $u$ in the class, there is a unique solution $\bar{x}$ that  is defined and bounded for all $k \in \mathbb{Z}$.

\item There exists a $\mathcal{KL}$ function $\beta$, independent of $u$ in the class, such that  
$
\|\overline{x}_k - x_k\| \leq \beta(\|\overline{x}_{k_0}-x_{k_0}\|,k-k_0),
$
for any initial time $k_0 \in \mathbb{Z}$, any $k \geq k_0$ and any initial state $x_{k_0}$.
\end{enumerate}
 \end{definition}

Let $\widetilde{\mathbb{V}}(X,\mathcal{D})$ denote the class of discrete-time signals $\widetilde{V}_{\cdot}$ on $X=\mathbb{Z}$ or $X=\mathbb{Z}_-$ that take values in a compact set $\mathcal{D}$, and let $\widetilde{w}_{\cdot}$ be a real weighting sequence  on $\mathbb{Z}_-$ such that $\widetilde{w}_k \in (0,1]$ for all $k \in \mathbb{Z}_-$ and $\mathop{\lim}_{k \rightarrow -\infty} \widetilde{w}_k=0$. Define the norm $\| \cdot \|_{\widetilde{w}}$ on $\widetilde{V}_{\cdot} \in \widetilde{\mathbb{V}}(\mathbb{Z}_-,\mathcal{D})$ by $\|\widetilde{V}\|_{\widetilde{w}}=\mathop{\sup}_{k \in \mathbb{Z}_-}|\widetilde{w}_k \widetilde{V}_k|$ and the signal class $\mathscr{S}(\mathbb{Z}_-,\mathcal{D})=\{\widetilde{V}_{\cdot} \in \widetilde{\mathbb{V}}(\mathbb{Z}_-,\mathcal{D})  \mid \| \widetilde{V} \|_{\widetilde{w}} < \infty\}$. The signal class $\mathscr{S}(\mathbb{Z}_-,\mathcal{D})$ together with the norm $\|\cdot\|_{\widetilde{w}}$ defines a metric space $(\mathscr{S}(\mathbb{Z}_-,\mathcal{D}),\|\cdot\|_{\widetilde{w}})$ with the metric $d(\widetilde{V},\widetilde{V}')=\|\widetilde{V}-\widetilde{V}'\|_{\widetilde{w}}$.

\begin{definition}[\cite{grigoryeva2018universal}] \label{def:DT-FM-functional}
A functional $F: \mathscr{S}(\mathbb{Z}_-,\mathcal{D}) \rightarrow \mathbb{R}$ is said to be a fading memory functional (with respect to the weighting sequence $\widetilde{w}$) if $F$ is a continuous functional  on $(\mathscr{S}(\mathbb{Z}_-,\mathcal{D}),\|\cdot\|_{\widetilde{w}})$. 
\end{definition}

The definition of a discrete-time fading memory filter is analogous to the continuous-time version given in Definition \ref{def:CT-FM-filter} but with the continuous-time fading memory functional being replaced with its discrete-time counterpart given in Definition \ref{def:DT-FM-functional}. Letting $\widetilde{\Pi}_0$ be the projection of a signal $\widetilde{V}: \mathbb{Z} \rightarrow \mathbb{R}$ onto its restriction  on $\mathbb{Z}_-$, $(\widetilde{\Pi}_0 \widetilde{V})_k = \widetilde{V}_k$ for all $k \in \mathbb{Z}_-$, the definition is given below:
\begin{definition}[\cite{grigoryeva2018universal}]
\label{def:DT-FM-filter}
A fading memory filter $U_F$ is a causal map from an input sequence  $\widetilde{V}_{\cdot} \in \widetilde{\mathbb{V}}(\mathbb{Z},\mathcal{D})$ to an output sequence $\widetilde{Y}_{\cdot}$ induced by a fading memory functional $F$ as defined by
\begin{equation}
\widetilde{Y}_k = U_F(\widetilde{V}_{\cdot})_k=F(\widetilde{\Pi}_0(\sigma_k(\widetilde{V}_{\cdot})).
\end{equation}
\end{definition}

The main result for the discrete-time DMS is given in the theorem below, with the proof given in the Methods section.

\begin{theorem}
\label{thm:DT-FM}
Let $\mathcal{D}=[a,b]$ with $-\infty<a<b<\infty$ and $\nu=\min_{v \in \mathcal{D}} K(v)$ ($K=-A$). The discrete-time system \eqref{eq:discrete-time} has the convergence property for any class of bounded input sequences and for any weighting sequence $\widetilde{w}$ satisfying
$$
\sum_{k=-\infty}^{-1} \frac{|k| e^{-(|k|-1)\nu T_s}}{\min\{\widetilde{w}_l\}_{k \leq l \leq 0}}<\infty
$$
and
$$
\sum_{k=-\infty}^0 \frac{e^{-|k|\nu T_s}}{\widetilde{w}_{k-1}} < \infty.
$$
In this case the dynamics induce the fading memory functional
\begin{align*}
F(\widetilde{V}_{\cdot}) &= \mathop{\lim}_{k_0 \rightarrow -\infty} \widetilde{P}^{AB}_{0} \\&= \sum_{k=-\infty}^{0} \left(\prod_{l=k}^{-1} \widetilde{\Phi}_l \right) k_{01}(\widetilde{V}_{k-1}) \int_{(k-1)T_s}^{kT_s} \Phi_{kT_s,\tau} d\tau
\end{align*}
on $\mathscr{S}(\mathbb{Z}_-,\mathcal{D})$, with the convention that $\prod_{l=0}^{-1} \widetilde{\Phi}_l = 1$, and the fading memory filter
\begin{align*}
U_F(\widetilde{V}_{\cdot})_k &= F(\widetilde{\Pi}_0\sigma_k(\widetilde{V}_{\cdot}))
\end{align*}
for any $\widetilde{V} \in \widetilde{\mathbb{V}}(\mathbb{Z}_-,\mathcal{D})$.
\end{theorem}

A weighting sequence that satisfies the requirement of Theorem \ref{thm:DT-FM} exists. For instance, it can be chosen to be
$\widetilde{w}_k = \max\{1,|k|\}e^{-a T_s |k|} $ for any $0< a < \nu$. In the discrete-time case it is known that if the input is uniformly bounded, as is the case here since $\widetilde{V}_k \in \mathcal{D}$ for all $k$, then if $F$ is continuous with respect to $\|\cdot\|_{\widetilde{w}}$ for some weighting sequence $\widetilde{w}$ then it is continuous for all weighting sequences \cite{grigoryeva2018echo}.  Thus the following holds.

\begin{corollary}
The functional $F$ in Theorem \ref{thm:DT-FM} has the fading memory property on  $(\mathscr{S}(\mathbb{Z}_-,\mathcal{D}),\|\cdot\|_{\widetilde{w}})$ for any weighting sequence $\widetilde{w}$.
\end{corollary}

\section{Discussion}
\label{sec:discussion} 
The purpose of this section is to present  and discuss the results of numerical simulations that illustrate features and behaviour of the DMS in continuous-time,  which are predicted by, and  consistent with, the results in \S \ref{sec:results}. The simulations are given for a DMS model with the parameter values given in Table \ref{tab:parameters}. The section then  summarises the contributions of the paper and discusses some directions for future work. 

\begin{figure}[h]
\centering
\includegraphics[scale=0.4]{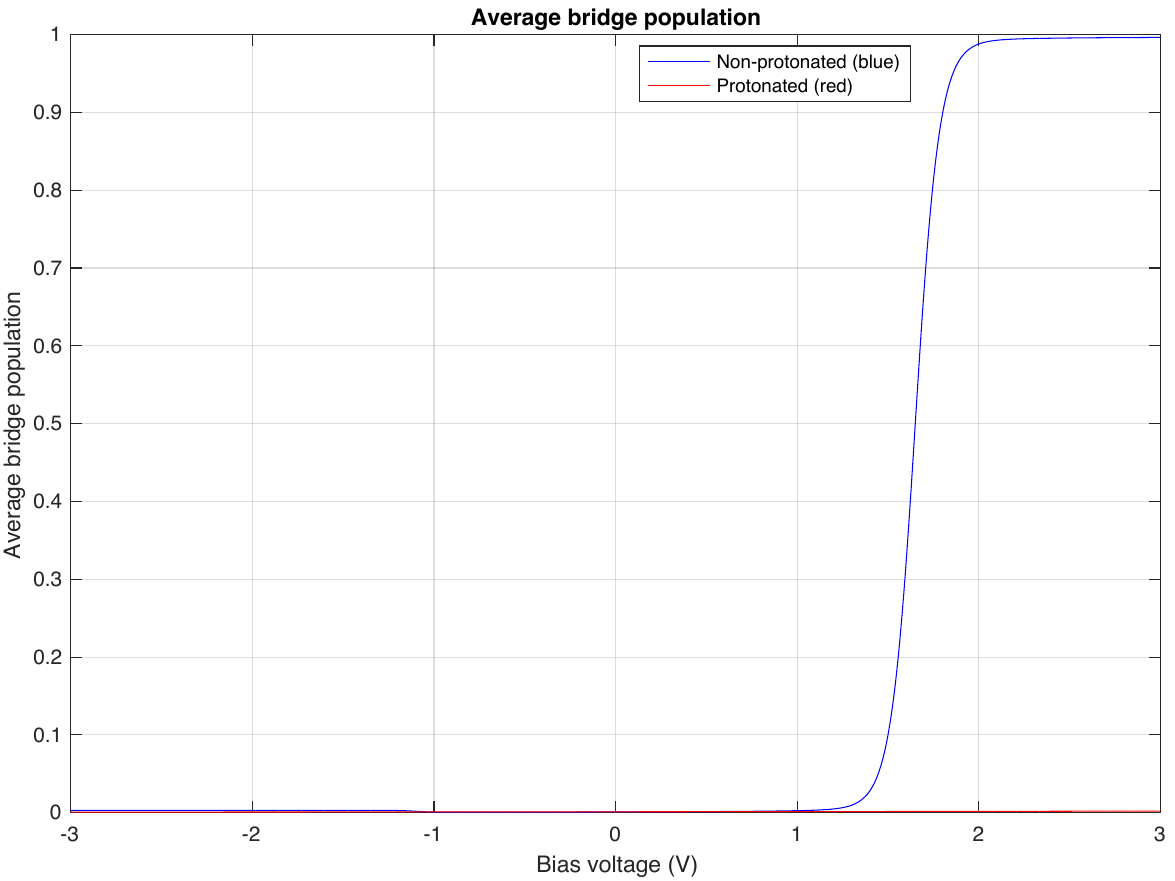}
\caption{The average bridge population as a function of bias voltage}  
\label{fig:average-bridge-pop}
\end{figure}

\begin{figure}[h]
\centering
\includegraphics[scale=0.4]{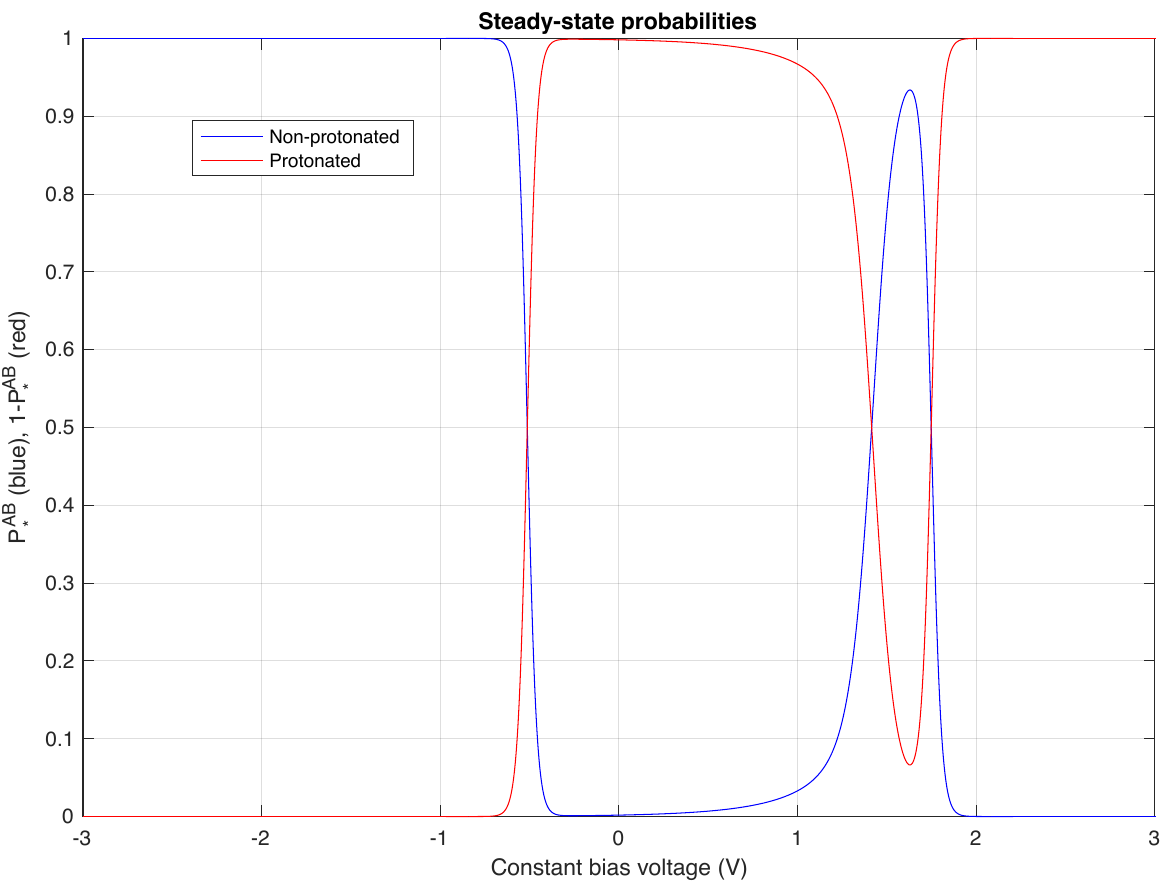}
\caption{The steady-state probability $P^{AB}_*$ against constant bias voltage values.}  
\label{fig:ss-fixed-bias}
\end{figure}

Fig.~\ref{fig:average-bridge-pop}  shows a plot of the average bridge population in the non-protonated ($\langle b_n \rangle^{AB}$, blue) and protonated $(\langle b_n \rangle^{\overline{AB}}$, red) state. In the protonated state the average bridge population is almost vanishing for all values of the bias voltage, while in the non-protonated state it takes on a value in the interval $[0,1]$, as stated in Lemma \ref{lem:positivity},  and takes on non-negligible values for sufficiently positive values of the bias voltage. 
Fig.~\ref{fig:ss-fixed-bias}  shows a plot of the steady-state probability $P^{AB}_*$  (probability of being in the non-protonated state) and $P^{\overline{AB}}_*=1-P^{AB}_*$ (probability of being in the protonated state) for various constant  bias voltage values, following Corollary \ref{cor:steady-state}. As shown in the plot $P^{AB}_*$ becomes negligible for sufficiently positive values of the bias voltage while for sufficiently negative values of the bias voltage it asymptotes to a constant value of 1, meaning that for the latter voltage values the DMS converges to the non-protonated state. $P^{AB}_*$ can also take on non-negligible values for bias voltages in a range between 0 and 2.  

\begin{figure}
\centering
\includegraphics[scale=0.32]{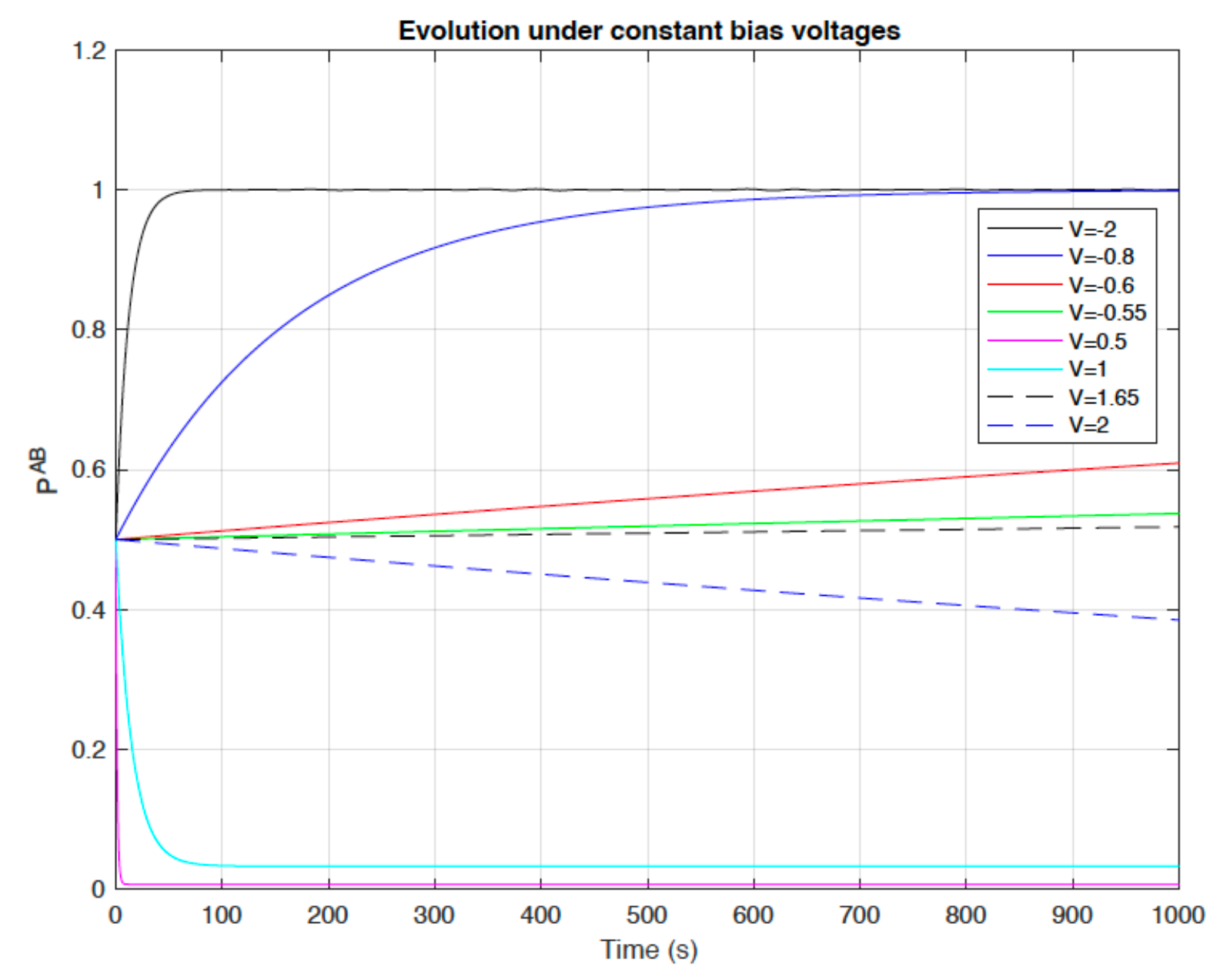}
\caption{Dynamic response of $P^{AB}_{\cdot}$ for different constant bias voltages ($v=-2,-0.8,-0.6,-0.55,0.5,1,1.65,2$ V).}  
\label{fig:response-fixed-bias}
\end{figure}

Recall that the transition function $\Phi_{t,t_0}$ satisfies $\Phi_{t,t_0} \leq e^{-\nu (t-t_0)}$ where $\nu = \min_{v \in \mathcal{D}} K(v)$ and $\nu >0$ since $K(v)>0$ for all $v \in \mathcal{D}$. Thus $\Phi_{t,t_0}$ goes to 0 as $t_0 \rightarrow -\infty$ at an exponential rate that depends on the actual bias voltage $V_{\cdot}$. The more positive the value of $K(V_t)$ over all $t$ the faster  $\Phi_{t,t_0}$ decays to 0. For a constant bias voltage $V_t=v$ for all $t$, $P^{AB}_t$ will converge to a constant steady-steady state value  $P^{AB}_*$ and the rate at which the former goes to the latter as $t \rightarrow \infty$ depends on the value of $v$. Fig.~\ref{fig:response-fixed-bias} shows the time evolution of $P^{AB}_{\cdot}$ for some choices of constant bias $v \in \{-2,-0.8,-0.6,-0.55,0.5,1,1.65,2\}$  from the initial state $P^{AB}_0=0.5$.  The associated steady-state value of $P^{AB}_{\cdot}$ for each constant bias voltage value can be surmised from the blue coloured plot in Fig.~\ref{fig:ss-fixed-bias}. 

\begin{figure}
\centering
\includegraphics[scale=0.43]{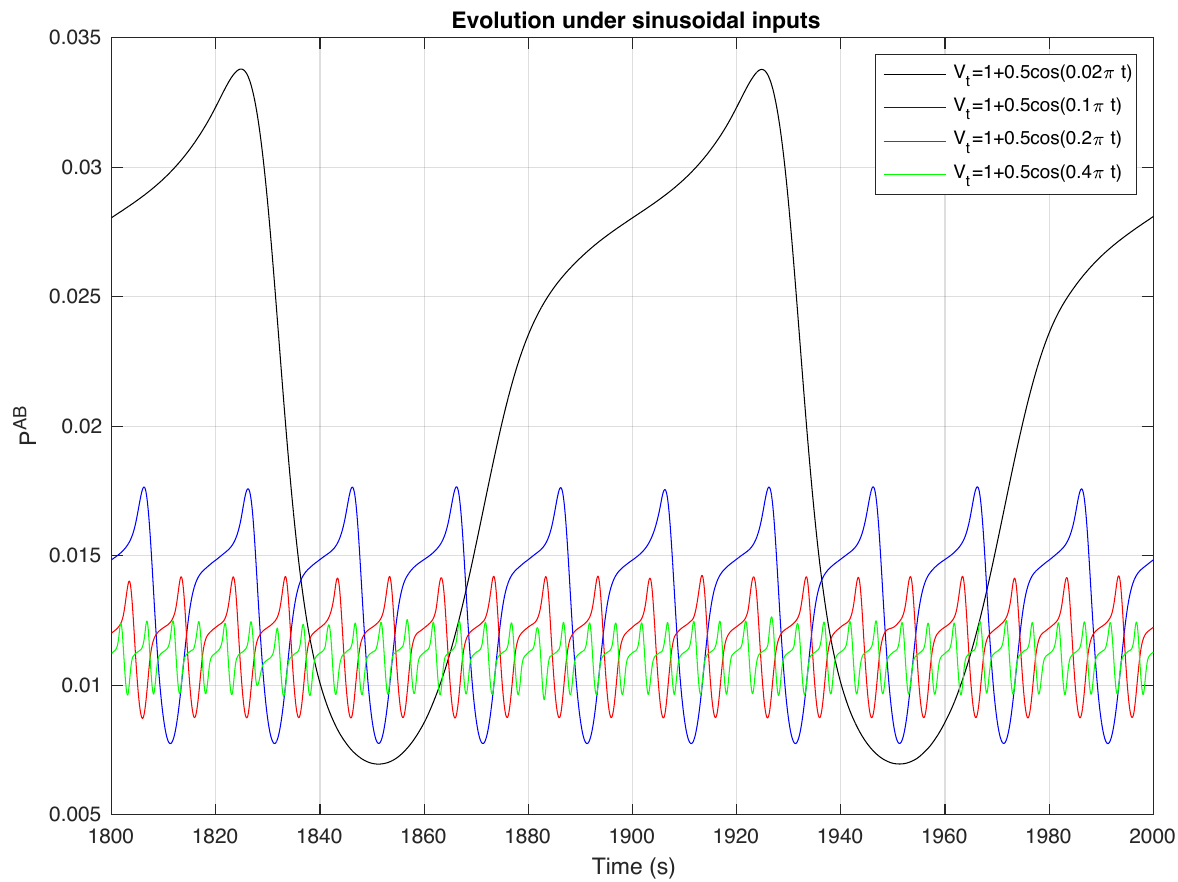}
\caption{Response of $P^{AB}$ to sinusoidal inputs}  
\label{fig:sinusoidal-probability}
\end{figure}

\begin{figure}
\centering
\includegraphics[scale=0.5]{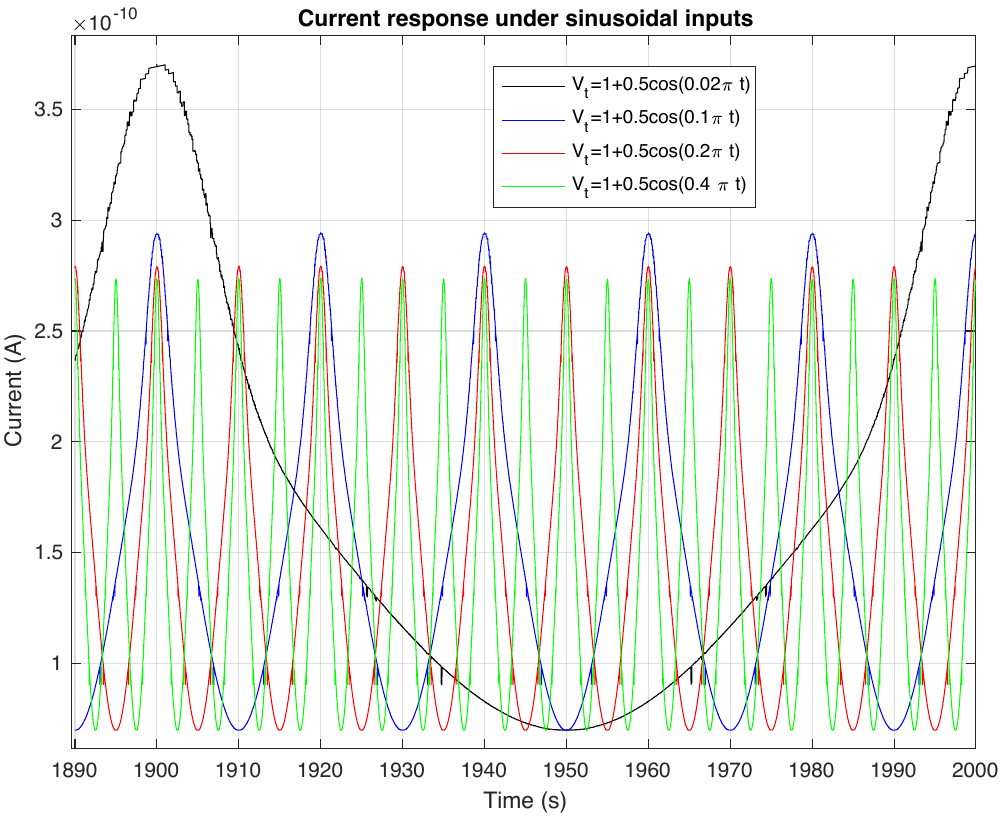}
\caption{Current response  to sinusoidal inputs}  
\label{fig:sinusoidal-current}
\end{figure}

Finally, for convergent dynamical systems, the response to a periodic input signal is again a periodic signal of the same period as the input \cite{PvdW16}. In the context of the DMS, given a periodic bias voltage $P^{AB}_{\cdot}$  the current response of the DMS will also be periodic. Fig.~\ref{fig:sinusoidal-probability} demonstrates the evolution of $P^{AB}$ for a  bias voltage of the form $V_t = 1 + 0.5 \cos(2 \pi f  t)$ for $f  \in \{0.01, 0.05,0.1,0.2\}$ Hz when the DMS is initialized at $P^{AB}_0=0$. The value of the constant bias 1 and the amplitude 0.5 of the periodic component were selected so that the DMS has a reasonable response time (cf. Fig.~\ref{fig:response-fixed-bias}) and the current responses lie  in a range where they are not always negligible (cf. Fig.~\ref{fig:DMS-current-response}). The current responses under the same sinusoidal inputs are shown in Fig.~\ref{fig:sinusoidal-current}. 
 
To conclude the discussion, in this paper we have rigorously established convergence and fading memory properties of a differential equation model inspired by the dynamic molecular switch in \cite{Wang22}. The model is distinguished in that it is linear-in-the-state, making it exactly solvable, but is nonlinear-in-the-input, and it exhibits both mimicry of biological synaptic behavior in the brain, which was previously demonstrated in experiments and in simulations using this model \cite{Wang22}.  At the same time the model possesses mathematical properties that enable stable learning of sequential data (in particular, forgetting the dependence on the initial condition of the model). To the best of our knowledge these features have not been {\em simultaneously} demonstrated in other synaptic models, which may lead to new links between learning processes in the brain and mathematical abstractions that underpin machine learning theory.  

The convergence and fading memory properties established for the model herein support the use of these switches in large-scale neuromorphic structures for temporal information processing.  The DMSs could be used as basic computational units that can be interconnected in deep layered structures akin to deep feedforward and recurrent neural networks, and variants thereof (e.g., LSTMs and GRUs). The stability properties of the DMS are expected to  be helpful in devising learning algorithms on these structures and establishing their convergence properties. Moreover, this work also points to the promise of general linear-in-the-state and nonlinear-in-the-input models for capturing synaptic behavior and stable processing of time-varying inputs.  For instance, beyond the DMS they may be useful as general parameterized models in data-driven modeling of arbitrary physical systems that exhibit these two features but for which a first principles model may be unknown or unavailable. These will be some of the topics for future research following from this work both from the side of algorithm development or mapping of the features onto other types of molecular hardware \cite{WLLJSYWG23,ZLTCGZWY23,KCSGvdB24,CHGHS25}.

\section{Methods}
\label{sec:methods}

In this Methods section we collect together the proofs for the results stated in the Results section.

\subsection{Proof of Lemma \ref{lem:positivity}}
Let $\gamma^{AB/\overline{AB}}>0$. Clearly, by definition, $\langle b_n^{AB/\overline{AB}} \rangle(v) \geq 0$ for all $v$.  
Then we find that, since $0 \leq f_{\pm}(E)(v) \leq 1$ for all $E$ and $v$, and $\gamma^{AB/\overline{AB}}=\gamma_L^{AB/\overline{AB}}+\gamma_R^{AB/\overline{AB}}$,
\begin{align*}
\langle b_n^{AB/\overline{AB}} \rangle(v) &\leq \frac{\gamma^{AB/\overline{AB}}}{2\pi} \int_{-\infty}^{\infty} \frac{1}{E^2 + (\gamma^{AB/\overline{AB}}/2)^2} dE \\
&= \frac{\gamma^{AB/\overline{AB}}}{2\pi} \frac{2}{\gamma^{AB/\overline{AB}}}\left. \arctan\left(\frac{2E}{\gamma^{AB/\overline{AB}}}\right) \right|_{E=-\infty}^{E=\infty}\\
&=\frac{1}{\pi} \pi\\
&=1.
\end{align*}
It follows from this that $k_{01}(v),k_{10}(v) \geq 0$ and $A(V) \leq 0$ for all $v$. Therefore, also $\Phi_{t,t_0} \geq 0$ for all $t,t_0$ with $t \geq t_0$, and $P^{AB}_{t} \geq 0$ for all $t \geq t_0$ when $P^{AB}_{t_0} \geq 0$ follows from \eqref{eq:P-solution}. 

Since $P^{\overline{AB}}_{\cdot} = 1 -P^{AB}_{\cdot}$, it follows from \eqref{eq:DMS-ODE-2} that it satisfies the ODE:
\begin{align*}
\dot{P}^{\overline{AB}}_t &= -k_{01}(V_t) + (k_{01}(V_t) + k_{10}(V_t)) P^{AB}_t \\
&= k_{10}(V_t) - (k_{10}(V_t) + k_{10}(V_t)) P^{\overline{AB}}_t.
\end{align*}
The solution is analogous to \eqref{eq:P-solution} by replacing $P^{AB}$ with $P^{\overline{AB}}$ and $k_{01}$ with $k_{10}$. It follows from the same argument as in the previous paragraph that $P^{\overline{AB}}_{t} \geq 0$ for all $t \geq t_0$ when $P^{\overline{AB}}_{t_0} \geq 0$.

\subsection{Proof of Corollary \ref{cor:proper-PAB}}

Since $P^{AB/\overline{AB}}_t \geq 0$ for all $t \geq t_0$ when $P^{AB/\overline{AB}}_{t_0} \geq 0$ by   Lemma \ref{lem:positivity}, from the identity $P^{AB}_t + P^{\overline{AB}}_t = 1$ for all $t$ it must therefore also  hold that $P^{AB/\overline{AB}}_t \leq 1$ for all $t \geq t_0$. 

\subsection{Proof of Lemma \ref{lem:stability}}

Since by definition $A(t) \leq 0$ for all $t$ and $\Phi_{t,t_0} \geq 0$, the following inequality holds:
\begin{align*}
\dot{\Phi}_{t,t_0} &= A(t) \Phi_{t,t_0} \\
&=-(k_{01}(V_t) + k_{10}(V_t))\Phi_{t,t_0},\\
&\leq - \left(\min_{v \in [a,b]} K(v) \right) \Phi_{t,t_0},
\end{align*}
where 
\begin{align}
K(v)= k_{01}(v) + k_{10}(v)=-A(v). \label{eq:def-K-fun}
\end{align}
Letting $\nu= \min_{v \in [a,b]} K(v)>0$, by Gronwall's Inequality it follows that 
$$
0 \leq \Phi_{t,t_0} \leq e^{-\nu (t-t_0)} \Phi_{t_0,t_0}= e^{-\nu (t-t_0)} .
$$
By the Pinching Theorem of calculus, it can be concluded that $\mathop{\lim}_{t_0 \rightarrow -\infty} \Phi_{t,t_0} =0$.

\subsection{Proof of Corollary \ref{cor:steady-state}}

The steady-state probability $P_*^{AB}$ for a constant bias voltage $v$ can be determined  by setting $\dot{P}_t^{AB}=0$. To this end,
$$
0=-P_*^{AB}\left( k_{01}(v) + k_{10}(v)\right) + k_{01}(v),
$$
and solving for $P_*^{AB}$ gives the desired expression for $P_*^{AB}$ and substituting this into \eqref{eq:junction-current} gives the expression for $I_*$. 

\subsection{Proof of Theorem \ref{thm:CT-FM}}

Let $P_{j,t}^{AB}$ be the solution of the ODE \eqref{eq:DMS-ODE-2} corresponding to two arbitrary initial conditions $P_{j,t_0}^{AB}$ ($j=1,2$) but driven  by the same bounded piece-wise continuous input $V_{\cdot}$ (i.e., $|V_t| \leq C$ for all $t$ for some $C>0$). Then from \eqref{eq:P-solution}:
$$
|P_{1,t}^{AB} - P_{2,t}^{AB}| =\Phi_{t,t_0} |P_{1,t_0}^{AB}-P_{2,t_0}^{AB}|,\;\forall t \geq t_0
$$
and therefore $\mathop{\lim}_{t_0 \rightarrow -\infty} |P_{1,t}^{AB}- P_{2,t}^{AB}|=0$. That is, the two solutions converge exponentially to one another at a rate of at least $\nu$, which depends on the particular input $V_{\cdot}$ (as discussed in \S \ref{subsec:asymptotic-ss}). This implies the existence of a unique fixed point $\overline{P}^{AB}_t$ given by 
$$
\overline{P}^{AB}_t = \mathop{\lim}_{t_0 
\rightarrow -\infty} P_{t}^{AB} = \int_{-\infty}^t \Phi_{t,\tau} k_{01}(V_{\tau})d\tau.
$$
The fixed point defines the unique solution of the DMS ODE for all $t \in \mathbb{R}$ and any bounded piece-wise continuous input $V_{\cdot}$.  All solutions of the DMS converge to this solution for any initial condition $P^{AB}_{t_0}$ as $t_0 \rightarrow -\infty$. Therefore, the DMS is convergent.

Define the functional $F$ induced by the DMS dynamics for each input $V \in \mathscr{S}(\mathbb{R}_-,\mathcal{D})$ by:
\begin{align*}
F(V_{\cdot}) &= \overline{P}^{AB}_0 \\
&= \int_{-\infty}^0 \Phi_{0,\tau} k_{01}(V_{\tau})d\tau.
\end{align*}
We will show that this functional has the fading memory property. For this write $\Phi_{0,\tau}$ as $\Phi_{0,\tau}(V_{0:\tau})$ to emphasize its dependence on $V_{0:\tau}$. Let $V$ and $V'$ be two signals in $\mathscr{S}(\mathbb{R}_-,\mathcal{D})$. It follows that 
\begin{align}
\lefteqn{|F(V)-F(V')|} \notag\\
& \leq \int_{-\infty}^{0} |\Phi_{0,\tau}(V_{0:\tau}) k_{01} (V_{\tau})-\Phi_{0,\tau}(V'_{0:\tau}) k_{01}(V'_{\tau})|d\tau, \notag\\
&\leq \int_{-\infty}^{0} \left(\vphantom{\int_{0}^t} \left|\Phi_{0,\tau}(V_{0:\tau})-\Phi_{0,\tau}(V'_{0:\tau})\right|k_{01}(V_{\tau})  \right. \notag \\
&\quad + \left. \vphantom{\int_{0}^t} \Phi(0,\tau)(V'_{0:\tau})\left|k_{01}(V_{\tau}) - k_{01}(V'_{\tau}) \right|\right) d\tau. \label{eq:ineq-F}
\end{align}
Since $k_{01}$ is by definition a continuous and differentiable function, by the Mean-Value Theorem of calculus, it holds that 
$k_{01}(V_t)-k_{01}(V'_t) = k'_{01}(c_t) (V_t-V'_t)$, where $k'_{01}$ is the derivative of $k_{01}$ with respect to its argument $v$ and $c_t$ is some value in the interval $[\min\{V_t,V'_t\},\max\{V_t,V'_t\}]$. 
By the definition of $\Phi_{0,\tau}(V_{0:\tau})$ and applying the mean value theorem it holds that
$$
\Phi_{0,\tau}(V_{0:\tau}) -  \Phi_{0,\tau}(V'_{0:\tau})= e^{d_{\tau}} \int_{\tau}^{0} (A(V_s)-A(V'_s))ds
$$ 
for some 
$$d_{\tau} \in [\min\{q(\tau,V),q(\tau,V')\},\max\{q(\tau,V),q(\tau,V')\}],$$
where $q(\tau,V) =\int_{\tau}^{0} A(V_s)ds$. By another application of the mean-value theorem, it also holds that
$$
A(V_s)-A(V'_s) = \left. \frac{d}{dv}A(v) \right|_{v=e_s} (V_s-V'_s),
$$
for some $e_s \in [\min\{V_s,V'_s\},\max\{V_s,V'_s\}]$. 

Now, since $V_t$ and $V'_t$ lie in the compact set $\mathcal{D}$, it follows that $|k'_{01}(c_t)| \leq \max_{v \in \mathcal{D}}|k'_{01}(v)|$. Also, recall that, since $A(v)<0$ for all $v$, $d_{\tau} \leq \nu \tau $ where $\nu=\min_{v \in \mathcal{D}}K(v)>0$. It thus follows that for $\tau <0$,
\begin{eqnarray*}
|k_{01}(V_t)-k_{01}(V'_t)| &\leq& g_1 |V_t-V'_t|\\
|\Phi_{0,\tau}(V_{0:\tau}) -  \Phi_{0,\tau}(V_{0:\tau})| &\leq& e^{\nu \tau}  g_2  \int_{\tau}^{0} |V_s-V'_s|ds,
\end{eqnarray*}
where $g_1=  \max_{v \in \mathcal{D}}|k'_{01}(v)|$ and $g_2=\max_{v \in \mathcal{D}} \left|\frac{\partial }{\partial v} A(v) \right|$. 
Consider the integral $H_1(V,V')=\int_{-\infty}^{0} \vphantom{\int_{0}^t} \left|\Phi_{0,\tau}(V_{0:\tau})-\Phi_{0,\tau}(V'_{0:\tau})\right|k_{10}(V_{\tau}) d\tau$. Using the bound for $|\Phi_{0,\tau}(V_{0:\tau}) -  \Phi_{0,\tau}(V_{0:\tau})|$ above, followed by making a change of coordinates $(\tau,s) \rightarrow (\tau',s')=(s,\tau)$ the integral can be bounded as
\begin{eqnarray*}
H_1(V,V') &\leq& g_2  \int_{-\infty}^{0} e^{\nu \tau} \int_{\tau}^0 |V_s-V'_s|ds d\tau \\
&= & g_2 \int_{-\infty}^{0} |V_s-V'_s|  \int_{-\infty}^s  e^{\nu \tau}
 d\tau ds\\
 &= & g_2 \int_{-\infty}^{0} w_{s}|V_s-V'_s|  \frac{\int_{-\infty}^s  e^{\nu \tau}
 d\tau}{w_s} ds,\\
&=& \frac{g_2}{\nu} \int_{-\infty}^{0} w_{s}|V_s-V'_s|  \frac{e^{\nu s}
}{w_s} ds.
\end{eqnarray*}
 If the weighting function $w_{\cdot}$ is such that
 $$
 \int_{-\infty}^0   \frac{  e^{\nu s}
 }{w_s} ds < \infty,
 $$
then it holds that
\begin{eqnarray*}
H_1(V,V') &\leq & \frac{g_2}{\nu}   \left( \int_{-\infty}^0 \frac{  e^{\nu s}
 }{w_s}ds  \right) \mathop{\sup}_{s\leq 0} (w_{s}|V_s-V'_s| )\\
&=& \frac{g_2}{\nu} \left(\int_{-\infty}^0  \frac{e^{\nu s}
 }{w_s}ds  \right) \|V-V'\|_{w}. 
\end{eqnarray*}
Consider now the integral $H_2(V,V')=\int_{-\infty}^0 \Phi_{0,\tau}(V'_{0:\tau})|k_{10}(V_{\tau})-k_{10}(V'_{\tau})|d\tau$. By the bound on $|k_{10}(V_{\tau})-k_{10}(V'_{\tau})|$ obtained above and since $\Phi_{0,\tau}(V'_{0:\tau}) \leq e^{\nu \tau}$, we have the bound:
\begin{eqnarray*}
H_2(V,V') &\leq& g_1 \int_{-\infty}^{0} e^{\nu \tau}  |V_{\tau}-V'_{\tau}|d\tau\\
&=& g_1 \int_{-\infty}^{0} \left( \frac{e^{\nu \tau}}{w_{\tau}}\right) w_{\tau} |V_{\tau}-V'_{\tau}|d\tau\\
&\leq& g_1 \left(\int_{-\infty}^{0}  \frac{e^{\nu \tau}}{w_{\tau}} d\tau \right)\mathop{\sup}_{\tau \leq 0} (w_{\tau}|V_{\tau}-V'_{\tau}|)\\
&=& g_1 \left(\int_{-\infty}^{0}  \frac{e^{\nu \tau}}{w_{\tau}} d\tau \right)\|V-V'\|_w
\end{eqnarray*}
Combining the two bounds on $H_1(V,V')$ and $H_2(V,V')$  yields the bound
\begin{align*}
|F(V)-F(V')|&\leq M \|V-V'\|_w,
\end{align*}
where
$$
M= (g_1 + g_2/\nu)\int_{-\infty}^{0}  \frac{e^{\nu \tau}}{w_{\tau}} d\tau.
$$
Therefore for any $\epsilon >0$ by setting $\|V-V'\|_w<
\delta$ with $0<\delta < \epsilon/M$, we have that $|F(V)-F(V')|<\epsilon$. In other words, $F$ is continuous on $(\mathscr{S}(\mathbb{R}_-,\mathcal{D}),\|\cdot\|_w)$. Therefore it is a fading memory functional. It follows immediately that $U_F$ as defined in the theorem is the associated fading memory filter.

\subsection{Proof of Theorem \ref{thm:DT-FM}}

Let $\widetilde{P}_{j,k}$ be the solution of \eqref{eq:discrete-time} corresponding to the initial condition  
$\widetilde{P}_{j,k_0}$ ($j=1,2$) for the same input sequence $\widetilde{V}_{\cdot}$. Note that for any $k \geq k_0$:
$$
|\widetilde{P}_{1,k}-\widetilde{P}_{2,k}| \leq \widetilde{\Phi}_{k-1} \cdots \widetilde{\Phi}_{k_0}  |\widetilde{P}_{1,k_0}-\widetilde{P}_{2,k_0}|,
$$
from which it follows that $\mathop{\lim}_{k_0 \rightarrow -\infty}|\widetilde{P}_{1,k}-\widetilde{P}_{2,k}| =0$ for any $k$ since $\widetilde{\Phi}_{k}\leq e^{-\nu T_s}<1$ for all $k$. This guarantees the convergence property of the system for any class of bounded input sequences with the $\mathcal{KL}$-function $\beta(x,k) = |x| e^{-\nu T_s k}$ and  the existence of a unique fixed point $\overline{P}^{AB}_k$, which is given by
\begin{align*}
\overline{P}^{AB}_k &=\mathop{\lim}_{k_0 \rightarrow -\infty} \widetilde{P}^{AB}_k\\
&= \sum_{m=-\infty}^{0} \left(\prod_{l=m}^{-1} \widetilde{\Phi}_{k+l} \right) k_{01}(\widetilde{V}_{k+m-1})  \int_{(k+m-1)T_s}^{(k+m)T_s} \Phi_{(k+m)T_s,\tau} d\tau,
\end{align*}
 with the convention that the term  $\prod_{l=m}^{-1} \widetilde{\Phi}_{k+l}$ is dropped (or set equal to 1) when $m=0$.
 
The fading memory functional $F$ associated with the discrete-time system is $F(\widetilde{V})= \overline{P}^{AB}_0$ as given in the theorem statement. For any two sequences $\widetilde{V},\widetilde{V}' \in \mathscr{S}(\mathbb{Z}_-,\mathcal{D})$, we have that
\begin{align*}
|F(\widetilde{V})-F(\widetilde{V}')|
&\leq \sum_{k=-\infty}^{0} \left| \left(\prod_{l=k}^{-1} \widetilde{\Phi}_l(\widetilde{V}_l) \right) k_{01}(\widetilde{V}_{k-1})   \int_{(k-1)T_s}^{kT_s} \Phi_{kT_s,\tau}(\widetilde{V}_{k-1}) d\tau \right.\\
&\quad    -\left(\prod_{l=k}^{-1} \widetilde{\Phi}_l(\widetilde{V}'_l) \right) k_{01}(\widetilde{V}'_{k-1})  \left. \int_{(k-1)T_s}^{kT_s} \Phi_{kT_s,\tau}(\widetilde{V}'_{k-1}) d\tau     \right|\\
&=\sum_{k=-\infty}^{0} \left( \left\{\sum_{l=k}^{-1} \left(\prod_{m=k}^{l-1}\widetilde{\Phi}_m(\widetilde{V}_m)\right)|\widetilde{\Phi}_l(\widetilde{V}_l)-\widetilde{\Phi}_l(\widetilde{V}'_l)| \right. \right.\\
&\qquad  \times \left(\prod_{m=l+1}^{-1}\widetilde{\Phi}_{m}(\widetilde{V}'_m)\right)k_{01}(\widetilde{V}_{k-1})  
 \left. \int_{(k-1)T_s}^{kT_s} \Phi_{kT_s,\tau}(\widetilde{V}_{k-1}) d\tau  \right\}  \\
&\qquad + \left(\prod_{l=k}^{-1}\widetilde{\Phi}_{l}(\widetilde{V}'_l)\right)\left| k_{01}(\widetilde{V}_{k-1})  \int_{(k-1)T_s}^{kT_s} \Phi_{kT_s,\tau}(\widetilde{V}_{k-1}) d\tau \right.\\
&\qquad \left. \left. - k_{01}(\widetilde{V}'_{k-1})   \int_{(k-1)T_s}^{kT_s} \Phi_{kT_s,\tau}(\widetilde{V}'_{k-1}) d\tau \right| \right),
\end{align*}
with the convention that sums or products with an upper index smaller than its lower index are dropped, and where $\Phi_{kT_s,\tau}(\widetilde{V}_{k})$ is just $\Phi_{kT_s,\tau}$ with its dependence on $\widetilde{V}_{k}$ made explicit. For the terms between the brackets $\{ \cdots \}$ in the sum over $k$, by the same argument using the Mean-Value Theorem in the proof of Theorem \ref{thm:CT-FM}, we can bound $|\widetilde{\Phi}_l(\widetilde{V}_l) -\widetilde{\Phi}_l(\widetilde{V}'_l)| \leq M | \widetilde{V}_l-\widetilde{V}'_l|$, where
$$
0< M = \max_{v \in \mathcal{D}}\left| \frac{d}{dv} e^{A(v)T_s}\right|<\infty. 
$$
Analogously define the function $G$ as
\begin{align*}
G(v) &= k_{01}(v)\int_{(k-1)T_s}^{kT_s} \Phi_{kT_s,\tau}(v)d\tau\\
&=k_{01}(v) \int_{(k-1)T_s}^{kT_s} e^{A(v)(kT_s-\tau)} d\tau \\
&= -k_{01}(v) \frac{1}{A(v)} \left[e^{A(v)(kT_s-\tau)} \right]_{\tau=(k-1)T_s}^{\tau=kT_s}\\
&= -\frac{k_{01}(v)}{A(v)} (1-e^{A(v)T_s}),
\end{align*}
where the last line is well-defined since $A(v)<0$ for all $v$. 

Repeating the same argument as above, we have the bound
$|G(\widetilde{V}_{k-1}) -G(\widetilde{V}'_{k-1})| \leq M' | \widetilde{V}_{k-1}-\widetilde{V}'_{k-1}|$, where
$$
0< M' = \max_{v \in \mathcal{D}}\left| \frac{d}{dv} G(v) \right|<\infty. 
$$
Also, the products $\left(\prod_{m=k}^{l-1}\widetilde{\Phi}_m(\widetilde{V}_m)\right)\left( \prod_{m=l+1}^{-1}\widetilde{\Phi}_{m}(\widetilde{V}'_m)\right)$ and $\prod_{l=k}^{-1}\widetilde{\Phi}_{l}(\widetilde{V}'_l)$ can be bounded from above by $e^{-(|k|-1) \nu  T_s}$ and $e^{-|k| \nu  T_s}$ for all $k<0$, respectively.

Putting all of the above together, we have the bound:
\begin{align*}
|F(\widetilde{V})-F(\widetilde{V}')| &\leq \sum_{k=-\infty}^{0} \left(\left\{ \sum_{l=k}^{-1} \frac{e^{-(|k|-1)\nu  T_s}}{\widetilde{w}_l} M \mathop{\max}_{v \in \mathcal{D}}G(v) \widetilde{w}_l |\widetilde{V}_{l}-\widetilde{V}'_{l}| \right\} \right. \\
&\quad  \left. \vphantom{\sum_{k=\infty}^{0}}  + e^{-|k|\nu T_s} M' |\widetilde{V}_{k-1}-\widetilde{V}'_{k-1}| \right)\\
&\leq \sum_{k=-\infty}^0 \left\{ M \mathop{\max}_{v \in \mathcal{D}}G(v) \frac{e^{-(|k|-1)\nu T_s}}{\min\{\widetilde{w}_l\}_{k \leq l \leq 0}} \sum_{l=k}^{-1} \widetilde{w}_l |\widetilde{V}_l -\widetilde{V}'_l| \right. \\
&\quad \left. + M' \frac{e^{-|k|\nu T_s}}{\widetilde{w}_{k-1}} \widetilde{w}_{k-1}|\widetilde{V}_{k-1}-\widetilde{V}'_{k-1}|\right\}\\
&\leq  M \mathop{\max}_{v \in \mathcal{D}}G(v) \|\widetilde{V} -\widetilde{V}'\|_{\widetilde{w}} \sum_{k=-\infty}^{-1}  \frac{|k| e^{-(|k|-1)\nu T_s}}{\min\{\widetilde{w}_l\}_{k \leq l \leq 0}}  \\
&\quad + M' \left(\sum_{k=-\infty}^0  \frac{e^{-|k|\nu T_s}}{\widetilde{w}_{k-1}} \right)\|\widetilde{V} -\widetilde{V}'\|_{\widetilde{w}}. 
\end{align*}
Let the weighting sequence $\widetilde{w}$ be  such that 
$$
c_1 =\sum_{k=-\infty}^{-1} \frac{|k|e^{-(|k|-1)\nu T_s}}{\min\{\widetilde{w}_l\}_{k \leq l \leq 0}}<\infty
$$
and
$$
c_2=\sum_{k=-\infty}^0  \frac{e^{-|k|\nu T_s}}{\widetilde{w}_{k-1}} < \infty.
$$
Then for any $\epsilon>0$ choosing $\widetilde{V}$ and $\widetilde{V}'$ such that
$$
\|\widetilde{V}-\widetilde{V}' \|_{\widetilde{w}} <\frac{\epsilon}{ M c_1 \max_{v \in \mathcal{D}} G(v) + M' c_2},
$$
gives $|F(\widetilde{V})-F(\widetilde{V}')|<\epsilon$. Therefore, $F$ is a continuous functional on the metric space $(\mathscr{S}(\mathbb{Z}_-,\mathcal{D}).\|\cdot \|_{\widetilde{w}})$. The associated fading memory functional $U_F$ then follows immediately from the definition of $F$.

\section*{Acknowledgments}
CN acknowledges support by the Dutch Research Council (NWO) under grant VI.C.222.037. HN and CN thanks Cameron Nickle and Enrique Del Barco for discussions on the DMS model developed in \cite{Wang22}. HN also thanks Cameron for generously sharing and explaining his Python code for simulating the model.

\bibliographystyle{IEEEtran}
\bibliography{refs}
\end{document}